\pdfoutput=1

\documentclass[11pt]{article}

\usepackage[preprint]{acl}

\usepackage{times}
\usepackage{latexsym}

\usepackage[T1]{fontenc}

\usepackage[utf8]{inputenc}

\usepackage{microtype}

\usepackage{inconsolata}

\usepackage{graphicx}
\usepackage{amsmath}

\title{Steps are all you need: Rethinking STEM Education with Prompt Engineering}

\author{Krishnasai Addala \\
  IIIT Delhi \\
  Delhi, India \\
  \scalebox{0.90}{\texttt{krishnasai20442@iiitd.ac.in}} \\\And
  Kabir Dev Paul Baghel \\
  IIIT Delhi \\
  Delhi, India \\
  \scalebox{0.90}{\texttt{kabir20564@iiitd.ac.in}} \\\And
  Navya Gupta \\
  MIDAS Lab \\
  Delhi, India \\
  \scalebox{0.90}{\texttt{guptanavya1808@gmail.com}} \\\AND
  Rishitej Reddy Vyalla \\
  IIIT Delhi \\
  Delhi, India \\
  \scalebox{0.90}{\texttt{rishitej23439@iiitd.ac.in}} \\\And
  Chhavi Kirtani \\
  IIIT Delhi \\
  Delhi, India \\
  \scalebox{0.90}{\texttt{chhavi18229@iiitd.ac.in}} \\\And
  Avinash Anand \\
  IIIT Delhi \\
  Delhi, India \\
  \scalebox{0.90}{\texttt{avinasha@iiitd.ac.in}} \\\AND
  Rajiv Ratn Shah \\
  IIIT Delhi \\
  Delhi, India \\
  \scalebox{0.90}{\texttt{rajivratn@iiitd.ac.in}} \\
  }

\begin{document}
\maketitle
\vspace*{2cm}
\begin{abstract}
Few shot and Chain-of-Thought prompting have shown promise when applied to Physics Question Answering Tasks, but are limited by the lack of mathematical ability inherent to LLMs, and are prone to hallucination. By utilizing a Mixture of Experts (MoE) Model, along with analogical prompting, we are able to show improved model performance when compared to the baseline on standard LLMs. We also survey the limits of these prompting techniques and the effects they have on model performance. 
Additionally, we propose Analogical CoT prompting, a prompting technique designed to allow smaller, open source models to leverage Analogical prompting, something they have struggled with, possibly due to a lack of specialist training data.
\end{abstract}






\maketitle

\section{Introduction}
The emergence of Large Language Models has transformed various industries. Significant advancements have been made in enhancing LLMs' capabilities to transform the education sector \cite{anand2023sciphyrag, anand2023revolutionizing, anand2024mathify, anand2024mm}. The ability of large language models to tailor learning experiences, meet individual student needs, and offer scalable educational solutions positions them as an ideal fit for academic purposes. This paper delves into the role of prompt engineering in enhancing high school STEM education.\newline
The educational landscape is increasingly seeking innovative and effective teaching approaches. Conventional teaching materials often fail to address the varied learning preferences and requirements of students. Different students learn in different ways and such personalized attention
\\ \\ \\ \\ \\ \\ \\
is not always available. This points to the necessity of a system which can coherently answer and explain questions to students. Recent advancements in natural language processing have significantly impacted  various domains, including education. Large language models have shown potential in enhancing educational tools and resources, particularly in STEM education. Previous research has demonstrated the effectiveness of LLMs in physics question answering \cite{anand2023sciphyrag} and citation generation \cite{anand2023context}. Building on these findings, this paper explores the use of prompt engineering to further improve the performance of LLMs in high school STEM education.\newline
Physics and Mathematics, core subjects within the high school curriculum, pose unique challenges. Its abstractions can be hard for both students and LLMs to follow, necessitating customized explanations for students to understand intricate theories. This is where the capabilities of large language models shine, offering assistance in digesting concepts and addressing queries, thereby enriching the educational process.\newline
In this study, we evaluated our prompts by using Mistral 7B and Mixtral 8x7B. The models were tested using the dataset published by \cite{anand2023revolutionizing}, along with StemStep, allowing us to assess a model's ability to solve STEM problems. LoRA\cite{hu2021lora} was employed to mitigate the computational challenges associated with inference. \newline
This work is structured as follows: Section \ref{literature} addresses the related work. In Section~\ref{dataset}, we discuss \textbf{StemStep}, our dataset contribution. In Section~\ref{experiments}, we outline our experimentation, and Section~\ref{results} shows the evaluation results. Section~\ref{discussion} discusses the analysis and study of this research, and the paper's limitations, future scope and conclusion are summarized in Section  \ref{limitations}, \ref{future scope} and \ref{conclusion}, respectively.

\section{Related Work\label{literature}}
\subsection{Chain of Thought}
Chain of Thought is a reasoning method that involves breaking down complex problems into simpler, more manageable parts, allowing an AI to "think" in a step-by-step manner similar to human problem-solving. Key studies in this area, such as by \cite{wei2023chainofthought}, have demonstrated how CoT can significantly improve the performance of AI models on complex tasks that require multi-step reasoning. Applications of CoT are widespread, including in natural language processing, where it enhances comprehension and problem-solving abilities of AI systems.
Chain of Thought is a cognitive-inspired approach that mimics human reasoning by breaking down complex problems into a series of intermediate steps or thoughts. This method enables AI models, particularly in the field of Natural Language Processing, to tackle tasks that require multi-step reasoning, such as mathematics problems, multi-hop question answering, and commonsense reasoning tasks.

\subsubsection{Key Studies and Findings}
Wei et al.\cite{wei2023chainofthought} demonstrated the effectiveness of CoT in improving language models' performance on complex reasoning tasks. 
The study found that prompting models to produce intermediate reasoning steps before reaching a final answer significantly improved problem-solving accuracy.
Chain of Thought Prompting in LLMs: Recent studies have investigated the application of Chain of Thought (CoT) prompting in large-scale language models like GPT-3. These studies have shown that even when not explicitly trained for CoT, large models can exhibit remarkable reasoning abilities when prompted correctly.
\subsubsection{Applications in AI}
\textbf{Educational Technologies:} CoT has been applied in educational software to provide step-by-step explanations to students, aiding in learning and understanding complex concepts.\newline
\textbf{Advanced Problem Solving:} In domains requiring complex decision-making, such as programming and engineering, CoT helps in breaking down tasks into more manageable sub-tasks, leading to more efficient and effective solutions.\newline
\textbf{Enhanced Comprehension in NLP:} By incorporating CoT, LLMs achieve a deeper understanding of text, improving performance in summarization, question answering, and text generation tasks by providing contextually rich, reasoned responses.

\subsubsection{Examples of CoT in Practice}
\leavevmode\newline 
\textbf{Math Word Problems:}
AI models use CoT to dissect math word problems into smaller arithmetic operations, sequentially solving each part to arrive at the final answer. This approach has the potential to overcome the difficulty most LLMs have with mathematical reasoning.\\
\textbf{Scientific Research:} 
LLMs have been shown to be helpful\cite{bryant2023assessing} in assisting researchers in Physics related research, with the caveat that they sometimes hallucinate in a way that disrupts their ability to respond properly.
CoT assists in parsing complex scientific texts, enabling models to outline step-by-step methodologies, results interpretation, and conclusions, thus aiding in literature review and hypothesis generation, all while reducing the hallucination rate\cite{Gupta2023InstructionTM} in responses.
\subsection{Few Shot Prompting}
Few Shot Prompting\cite{brown2020language} refers to the technique of training AI models with a very limited amount of data, typically just a few examples or "shots". This approach contrasts with Zero-Shot Learning, where the model receives no examples, and Many-Shot Learning, where the model is trained on large datasets. \newline
Recent advances in this area, particularly in the context of LLMs like GPT4\cite{openai2023gpt4}, have shown remarkable abilities to generalize from minimal data, making it a valuable technique for tasks where data is scarce or costly to obtain.
\subsection{Analogical Prompting}
Analogical Prompting, as framed in \cite{yasunaga2023large}, represents an innovative paradigm wherein large language models are endowed with the capacity for analogical reasoning. This methodology asks the model to generate contextually relevant exemplars, thereby augmenting the models' efficacy in navigating a spectrum of reasoning tasks. Distinct from conventional methodologies that necessitate manual exemplar curation, this approach heralds a more dynamic and efficient mechanism, demonstrably surpassing traditional 0-shot and few-shot prompting modalities in its applicability and outcomes. However, analogical prompting has been mainly tested on proprietary models such as GPT4, we investigate its efficacy on non-finetuned open source models such as Mistral 7B\cite{jiang2023mistral} and Mixtral 8x7B\cite{jiang2024mixtral}.
\subsection{Mixture of Experts}
The Mixture of Experts\cite{chen2022understanding} (MoE) model is a machine learning paradigm where multiple specialist models (experts) are trained on different parts of a problem and a gating mechanism decides which expert to use for a given input. This approach allows for more scalable and efficient learning, as each expert can become highly specialized. MoE has been successfully integrated with other AI techniques, such as deep learning and reinforcement learning, leading to significant improvements in tasks like language translation and image recognition.\newline


\subsection{Physics Question Answering using RAG}
SciPhyRAG, developed by \cite{anand2023sciphyrag}, is a retrieval-augmented model designed to improve the performance of LLMs in answering physics questions. Traditional LLMs often struggle with the complex and nuanced nature of physics problems, which require not only an understanding of fundamental concepts but also the ability to apply these concepts to solve specific problems. SciPhyRAG addresses these challenges by integrating external knowledge sources, such as scientific literature and databases, to provide more contextually relevant and accurate answers. This approach significantly enhances the model's ability to comprehend and apply intricate physics concepts, thereby improving the overall reliability and accuracy of its responses.

The SciPhyRAG model operates by first retrieving relevant information from external sources based on the input question. This retrieved information is then used to augment the LLM's internal knowledge, allowing it to generate more informed and precise answers. By combining retrieval-augmented generation with advanced natural language processing techniques, SciPhyRAG sets a new standard for AI-driven physics education and research, demonstrating the potential for AI to assist in fields that require deep, domain-specific knowledge.\newline

\subsection{Advancements in Controllable Text Generation}
Goel\cite{goel2023advancements} explored significant advancements in the methods for controllable text generation, particularly focusing on the fine-tuning of LLMs for specific scientific domains. The study highlighted the importance of controlling various aspects of text generation, such as tone, style, and content relevance, to produce high-quality, domain-specific content. This capability is particularly crucial in educational applications, where the precision and clarity of generated content can directly impact learning outcomes.

Goel's work involved developing techniques to enable LLMs to generate text that adheres to specific constraints and guidelines provided by the user. By fine-tuning models with data from specialized scientific domains, the research demonstrated that LLMs could produce more accurate and contextually appropriate content. This advancement not only improves the usability of LLMs in academic and professional settings but also enhances their potential as tools for personalized education and automated content creation in various fields.\newline

\subsection{Citation Generation}
The work on context-enhanced language models by \cite{anand2023context} and the KG-CTG model \cite{anand2023kg} demonstrates the effectiveness of using knowledge graphs to guide LLMs in generating precise citations. These approaches improve the accuracy and relevance of citations in scholarly papers, showcasing the potential of LLMs in academic writing.\newline

\subsection{Multimodal Physics Question Answering}
MM-PhyQA\cite{anand2024mm}, a multimodal question-answering system that uses multiple images for chain-of-thought prompting. This approach enhances the accuracy and depth of answers in physics education by leveraging visual aids alongside textual explanations.\newline

\subsection{Mathematical Problem Solving}
Mathify\cite{anand23mathify} assessed the performance of LLMs on mathematical problem-solving tasks. The results reveal significant challenges that LLMs encounter with complex mathematical queries, emphasizing the necessity for specialized training data to enhance their problem-solving abilities.

\subsection{Similar Datasets}
\subsubsection{\textbf{SciQ}\cite{welbl2017crowdsourcing}} SciQ comprises 13.7K science-based questions, but these are multiple-choice format. In contrast, the dataset we present mainly consists of questions that demand several intermediate steps to solve, making them better suited for a CoT paradigm. Furthermore, we include mathematics questions to ensure a more diverse and balanced dataset.
\subsubsection{\textbf{GSM8K}\cite{cobbe2021training}}
GSM8K contains questions that are of middle school difficulty, which is below the standard required to evaluate prompts aimed at solving high school level problems.
\subsubsection{\textbf{JEEBench}\cite{arora2023have}}
JEEBench consists of approximately 450 questions in the domain of pre-engineering studies. These questions are generally more challenging than typical high school STEM problems and are primarily in multiple-choice format, which is often less suitable for prompting or training applications.
\section{Dataset\label{dataset}}
We introduce \textbf{StemStep} as a dataset that can be used to evaluate models in answering both physics and mathematics questions at a high school level.
The dataset contains 928 high-quality questions on all topics relevant to high school physics and mathematics.
Each question in the dataset is accompanied by a set of steps required to reach the final answer.
Below is a sample question from the dataset.\newline \newline
\textbf{Question:} "100g  of water is supercooled to -10 degrees . At this point, due to some disturbance , mechanised or otherwise. Some of it suddenly freezes to ice. What will be the temperature of the resultant mixture and how much mass would freeze?\\ \textbf{Steps:}" "Mass of water =100g ",\newline
 "At -10 degrees  the mixture has water and ice",\newline
 "Heat required by the mixture is   =(100)(1)(0-(-10)=1000 Cal", \newline
"Therefore, the mass of the mixture, =12.5g "\\ \textbf{Final answer:} Therefore, the mass of the mixture =12.5g.\newline
\newline
The data was gathered by scraping online resources aimed at high school students studying mathematics and physics. The data was scraped as LaTex text in order to maintain semantic integrity of the answers and their explanations.\newline

\begin{figure}
    \centering
    \includegraphics[width=1\linewidth]{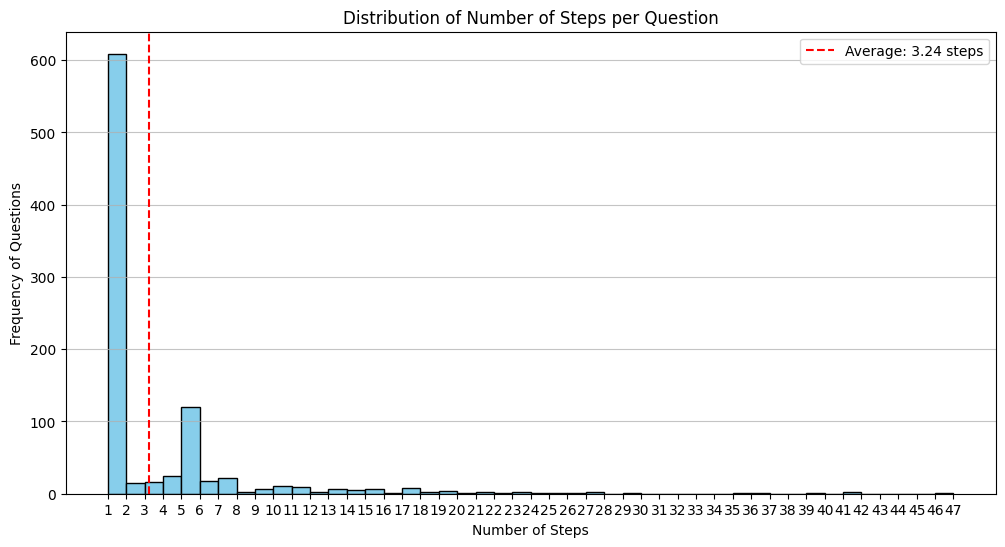}
    \caption{The Average number of Steps per Question is 3.24}
    \label{fig:avgnumstep}
\end{figure}

\textbf{Figure \ref{fig:avgnumstep}} shows the distribution of the number of steps for a given question in the dataset. \newline

\begin{figure}
        \centering
\includegraphics[width=1\linewidth]{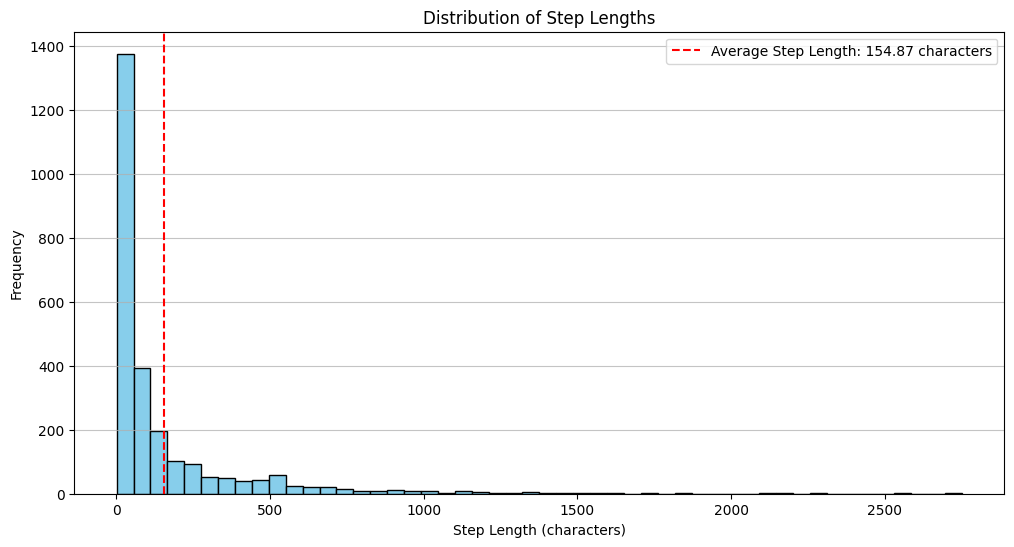}
        
    \caption{The Average Step Length is 154.9 Characters}
    \label{fig:avgsteplen}
\end{figure}

\textbf{Figure \ref{fig:avgsteplen}} shows the step length per question distribution throughout the dataset.\newline
\newline
It is clear to see that the dataset is biased towards questions with shorter explanations, which should be easier for models to comprehend.
\subsection{Student survey}
We verified the quality of the dataset by asking 5 individuals familiar with high-school level Physics and Mathematics to rate a sample of data points out of 10. These evaluators cumulatively gave the dataset an average score of $\frac{9}{10}$ after reviewing 50 questions each, indicating a \textbf{strong alignment} between the data and the human student's expectation of what would be helpful in learning the given subjects.

\section{Experimentation\label{experiments}}
We frame the objective of answering Physics questions as a text-generation task in order to leverage the strengths of smaller models which may struggle with heavier reasoning tasks.
We use the models Mistral 7B and Mixtral 8x7B from mistralai, in order to investigate to what extent an MoE model affects the performance on these tasks.
We establish a baseline for each model by asking the model to solve a given problem as a 0-Shot\cite{kojima2023large} prompt. The presence of the baseline should serve to contextualize the expected performance gains from prompt engineering.
\subsection{K Variable K Shot Prompting} 
We begin by investigating the effect of the number of examples in a few shot prompt on model performance, i.e, the effect of varying K in a K-shot prompt.
We construct the K-Shot CoT prompts by following the pipeline proposed by \cite{anand2023revolutionizing} 
, however, we utilize the steps from our dataset to guide the model into generating better structured answers.
\newline
We ran inference on base Mistral 7B, for its effectiveness in Text Generation and Question Answering tasks. We penalized the model for repetitive answers and quantized Mistral using LoRA to reduce hardware requirements.\newline
\subsubsection{\textbf{Example 1: Projectile Motion-:}}

\textbf{Question:}  
"A ball is thrown vertically upwards with an initial velocity of 20 m/s. Calculate the maximum height reached by the ball. Assume \( g = 9.8 \, m/s^2 \)."\newline

\textbf{3-Shot CoT Prompt:}  
1. "A stone is thrown vertically upwards with an initial velocity of 15 m/s. Calculate the maximum height reached by the stone."  \newline
2. "An arrow is shot vertically upwards with an initial velocity of 25 m/s. Calculate the maximum height reached by the arrow."  \newline
3. "A rocket is launched vertically upwards with an initial velocity of 30 m/s. Calculate the maximum height reached by the rocket." \newline

\textbf{Model Response:}  
Given:
\begin{itemize}
    \item Initial velocity, \( u = 20 \, m/s \)
    \item Acceleration due to gravity, \( g = 9.8 \, m/s^2 \)
\end{itemize}

Using the equation of motion:
\[ v^2 = u^2 - 2gh \]

At maximum height, the final velocity \( v = 0 \):
\[ 0 = (20)^2 - 2 \times 9.8 \times h \]
\[ 400 = 19.6h \]
\[ h = \frac{400}{19.6} \]
\[ h = 20.41 \, m \]

\subsubsection{\textbf{Example 2: Electric Field-:}}

\textbf{Question:}  \newline
"Calculate the electric field at a point 0.5 m away from a charge of \( 2 \times 10^{-6} \, C \). Use Coulomb's constant \( k = 8.99 \times 10^9 \, Nm^2/C^2 \)."\newline

\textbf{3-Shot CoT Prompt:}  
1. Calculate the electric field at a point 1 m away from a charge of \( 1 \times 10^{-6} \, C \). \newline 
2. Calculate the electric field at a point 0.2 m away from a charge of \( 3 \times 10^{-6} \, C \). \newline
3. Calculate the electric field at a point 0.8 m away from a charge of \( 4 \times 10^{-6} \, C \). \newline

\textbf{Model Response:}  
Given:
\begin{itemize}
    \item Charge, \( Q = 2 \times 10^{-6} \, C \)
    \item Distance, \( r = 0.5 \, m \)
    \item Coulomb's constant, \( k = 8.99 \times 10^9 \, Nm^2/C^2 \)
\end{itemize}

Using Coulomb's law:
\[ E = \frac{kQ}{r^2} \]
\[ E = \frac{8.99 \times 10^9 \times 2 \times 10^{-6}}{(0.5)^2} \]
\[ E = \frac{8.99 \times 10^9 \times 2 \times 10^{-6}}{0.25} \]
\[ E = \frac{17.98 \times 10^3}{0.25} \]
\[ E = 7.19 \times 10^4 \, N/C \]

The results of this study can be seen in Figure \ref{fig:image1}, \ref{fig:comparison}.\newline
The accuracy difference when using Mixtral 8x7B is shown in Table \ref{tab:accuracy_comparison}. 
We did not observe a large improvement in performance, although we did observe a moderate increase in accuracy.
\subsection{Analogical Prompting}
We adapt Analogical prompting from \cite{yasunaga2023large} in order to investigate its utility in this domain.
We asked the models to self-supply varying numbers of examples and measured the performance of the model in each scenario.\newline
In practice, we observe that the base models are unable to leverage analogical prompting, as they do not properly recall examples. This may be due to the \textbf{relatively small datasets} these models are trained on compared to models such as GPT3.5 or GPT4 which are able to recall relevant examples and solve the physics problems.
Both Mistral 7B and Mixtral 8x7B give qualitatively poor answers, and more often than not do not address the given question.\newline
\subsubsection{\textbf{Example 3: Analogical Prompting for Kinematics-:}}

\textbf{Question:}  
"A car accelerates from rest at a constant rate of \( 2 \, m/s^2 \). Calculate the time it takes to reach a velocity of \( 20 \, m/s \)."\newline

\textbf{Analogical Prompt:}  \newline
Recall relevant exemplars and solve the question: "A car accelerates from rest at a constant rate of \( 2 \, m/s^2 \). Calculate the time it takes to reach a velocity of \( 20 \, m/s \)."  \newline

\textbf{Model Response:}  
Given:
\begin{itemize}
    \item Initial velocity, \( u = 0 \)
    \item Acceleration, \( a = 2 \, m/s^2 \)
    \item Final velocity, \( v = 20 \, m/s \)
\end{itemize}

Using the equation of motion:
\[ v = u + at \]
\[ 20 = 0 + 2t \]
\[ t = \frac{20}{2} \]
\[ t = 10 \, s \]

\subsection{Analogical CoT Prompting}
In order to allow open-source models to leverage the strengths of both Few Shot CoT and Analogical prompting, we propose combining the two to guide the model in the kind of self-supplied examples it should generate.
We supply the model with 3 examples, and ask it to generate 3 further examples on its own. 
These values were chosen on analysis of prior work\cite{yasunaga2023large} done in this domain.\newline
\newline
We tested the prompt on GPT4 and Gemini Pro\cite{geminiteam2023gemini}, in addition to the Mistral models to gauge prompt viability.
\subsubsection{\textbf{Example 4: Combined CoT and Analogical Prompting for Energy-:}}

\textbf{Question:}  
"A 5 kg block slides down a frictionless inclined plane with an angle of 30 degrees. Calculate the speed of the block after sliding 2 meters."\newline

\textbf{3-Shot CoT + Analogical Prompt:}\newline  
1. A 3 kg block slides down a frictionless inclined plane with an angle of 45 degrees. Calculate the speed of the block after sliding 1 meter.  \newline 
2. A 4 kg block slides down a frictionless inclined plane with an angle of 60 degrees. Calculate the speed of the block after sliding 1.5 meters.  \newline 
3. A 6 kg block slides down a frictionless inclined plane with an angle of 30 degrees. Calculate the speed of the block after sliding 2.5 meters. \newline
Recall relevant exemplars and solve the question: "A 5 kg block slides down a frictionless inclined plane with an angle of 30 degrees. Calculate the speed of the block after sliding 2 meters."\newline
\textbf{Model Response:}  
Given:
\begin{itemize}
    \item Mass, \( m = 5 \, kg \)
    \item Angle, \( \theta = 30^\circ \)
    \item Distance, \( d = 2 \, m \)
    \item Acceleration due to gravity, \( g = 9.8 \, m/s^2 \)
\end{itemize}

The component of gravitational force along the incline:
\[ F = mg \sin(\theta) \]
\[ F = 5 \times 9.8 \times \sin(30^\circ) \]
\[ F = 5 \times 9.8 \times 0.5 \]
\[ F = 24.5 \, N \]

Using the work-energy principle:
\[ W = F \times d \]
\[ W = 24.5 \times 2 \]
\[ W = 49 \, J \]

Since \( W = \Delta K \):
\[ \Delta K = \frac{1}{2}mv^2 \]
\[ 49 = \frac{1}{2} \times 5 \times v^2 \]
\[ 49 = 2.5v^2 \]
\[ v^2 = \frac{49}{2.5} \]
\[ v = \sqrt{19.6} \]
\[ v = 4.43 \, m/s \]

We observe that GPT4 and Gemini Pro are able to give qualitatively better answers, and are able to answer more accurately than a simple CoT prompt. Mistral 7B struggled to generate relevant examples on its own, however it was able to answer the questions to a reasonable degree of accuracy, keeping in mind the lower parameter count and quantized nature.\newline
Of the open-source models, only Mixtral 8x7B was able to succesfully answer questions using this prompting style at a better than chance accuracy, Mistral 7B struggled to generate relevant text.\newline
The results of this study can be seen in Figure \ref{fig:comparison1}.

\section{Results\label{results}}

    \textbf{Mistral 7B:}
    We observe that the model performance peaks around K=3, while higher levels of K show that the performance slightly decreased and eventually plateaued.
    The metrics recorded signify that the model is able to generate text that is relatively semantically similar to the human-produced ground truth.
    The model also produced qualitatively good responses relative to the baseline and did not hallucinate when given some direction.\newline
 \textbf{However,}
        this model was unable to leverage Analogical prompting at any real level, perhaps due to the small parameter count or due to the lack of specialized domain specific data in the training stage. \newline\newline
 \textbf{Mixtral 8x7B:}
     The model did not display a discernible trend as Mistral 7B did, although K=6 seems to show the most consistent performance. This may be due to the larger context window being more accommodating of the longer prompts.\newline
    The variability of the recorded metrics show that the model had trouble consistently generating text that was very similar to the ground truth, however, the addition of CoT+Analogical prompting did lead to a more aligned answer relative to the baseline.
    Model accuracy did increase as more examples were added, upto K=6.
    The addition of pure Analogical prompting did not show a positive effect on accuracy, as the model had trouble answering the original question.\newline
\textbf{However:}
         This model was unable to leverage Analogical prompting in that it was unable to recall relevant examples, this may be addressed in the future by adding specialized domain specific data in the training stage.
\subsection{K-Shot CoT Performance}

The sharp decline in performance at K=8 highlights the limitations of the models' context window. Mixtral 8x7B consistently outperforms Mistral 7B, likely due to its mixture of experts architecture, which enhances its ability to leverage context effectively.

\subsection{Analogical Prompting Performance}

Mixtral 8x7B's superior performance in analogical prompting suggests that its architecture better supports the integration of analogical examples into its reasoning process. This could be due to its higher parameter count and more sophisticated attention mechanisms.

\subsection{Combined CoT and Analogical Prompting}

The combined approach significantly improves performance, particularly in more complex questions. This suggests that guiding the model in generating its own examples helps it better understand and solve the problems. The effectiveness of this approach is most evident in the physics domain, where combining conceptual understanding with procedural steps is crucial.
\subsection{Metrics\label{metrics}}
\begin{figure}[!htbp]
\centering
\includegraphics[width=\linewidth]{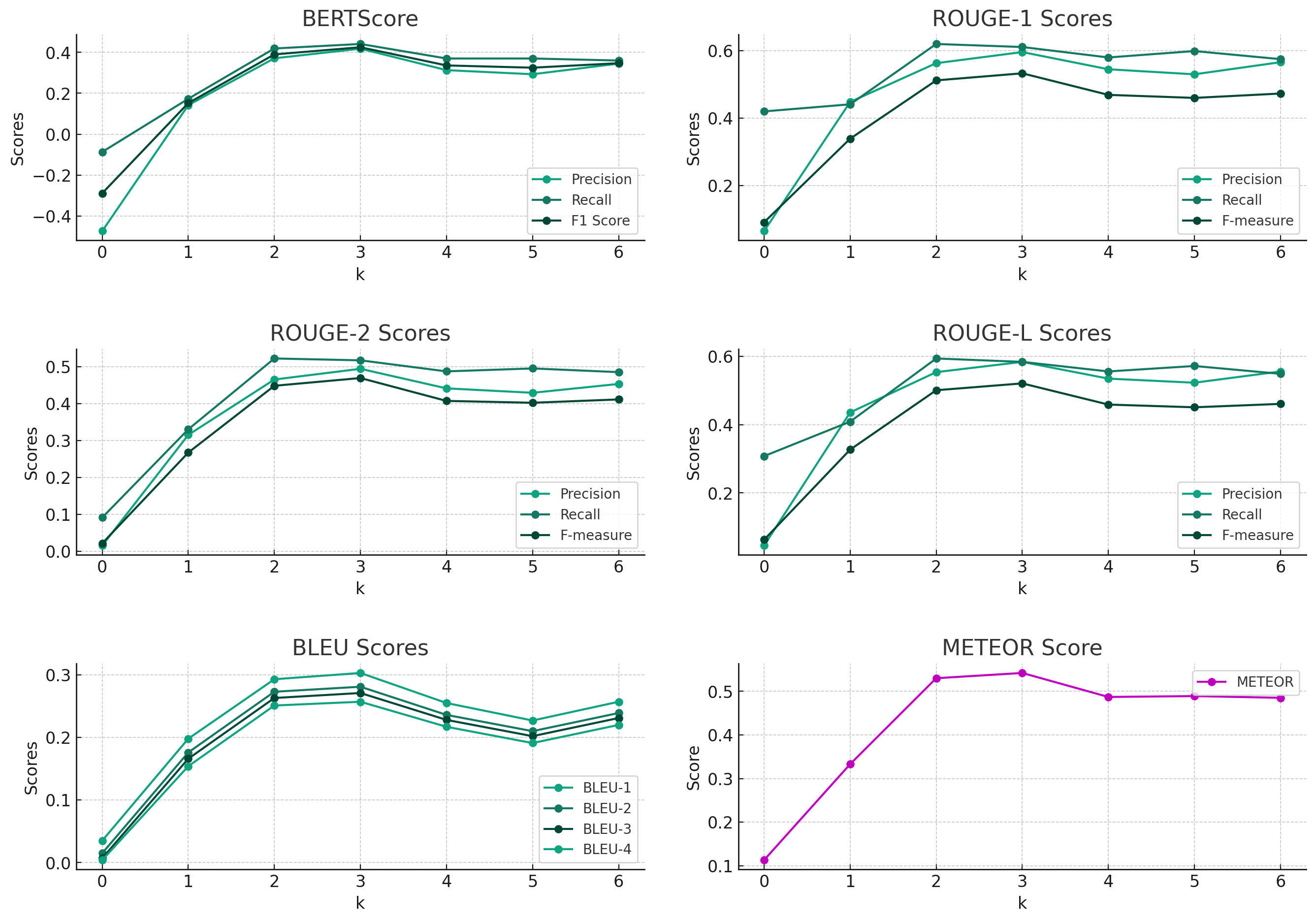}
\caption{K Variable K Shot Prompting on Mistral 7B}
\label{fig:image1}
\end{figure}

\begin{figure}[!hbp]
  \centering
  \includegraphics[width=\linewidth]{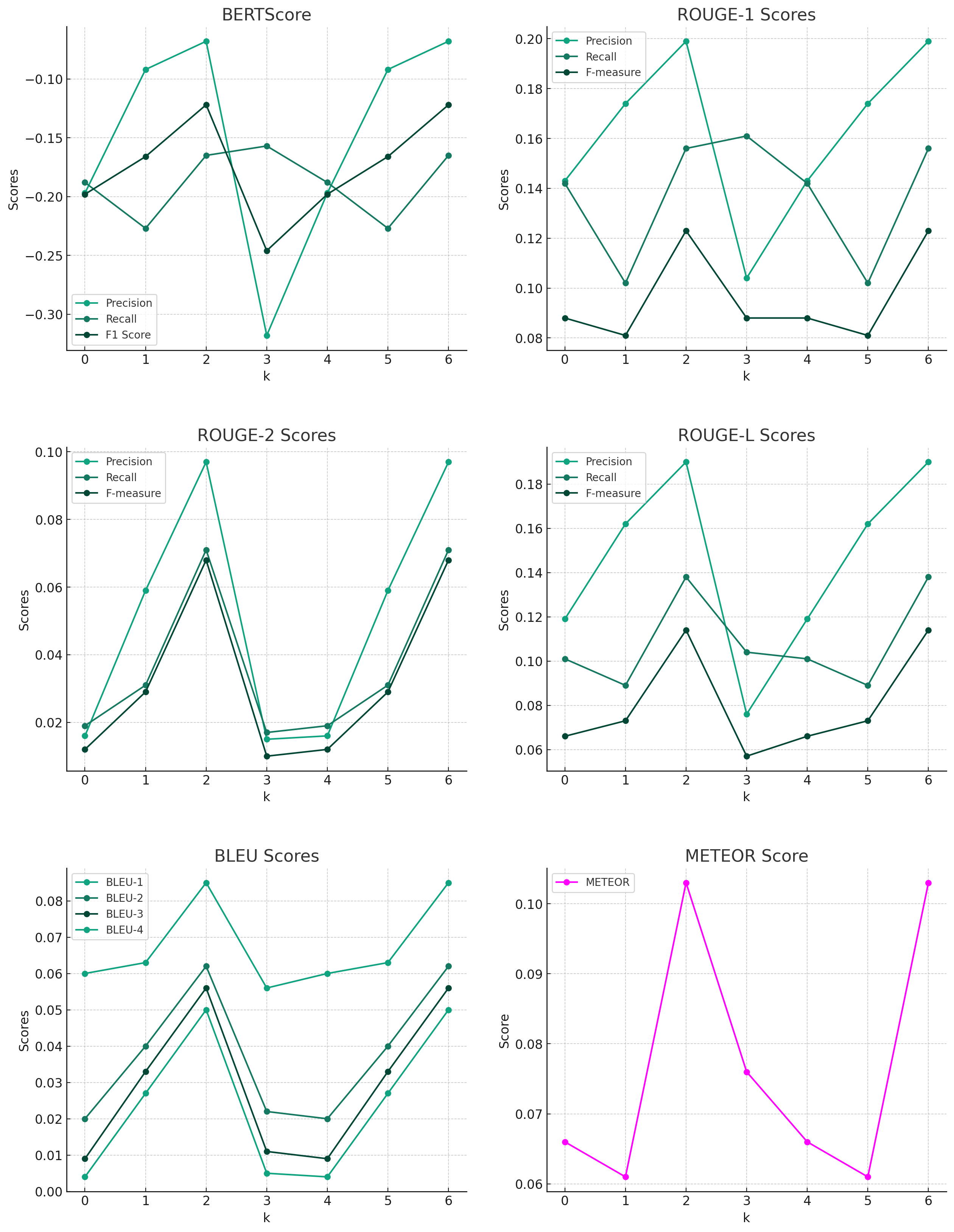} 
  \caption{K Variable K Shot Prompting on Mixtral 8x7B}
  \label{fig:comparison}
\end{figure}

\begin{figure}[!h]
  \centering
  \includegraphics[width=\linewidth]{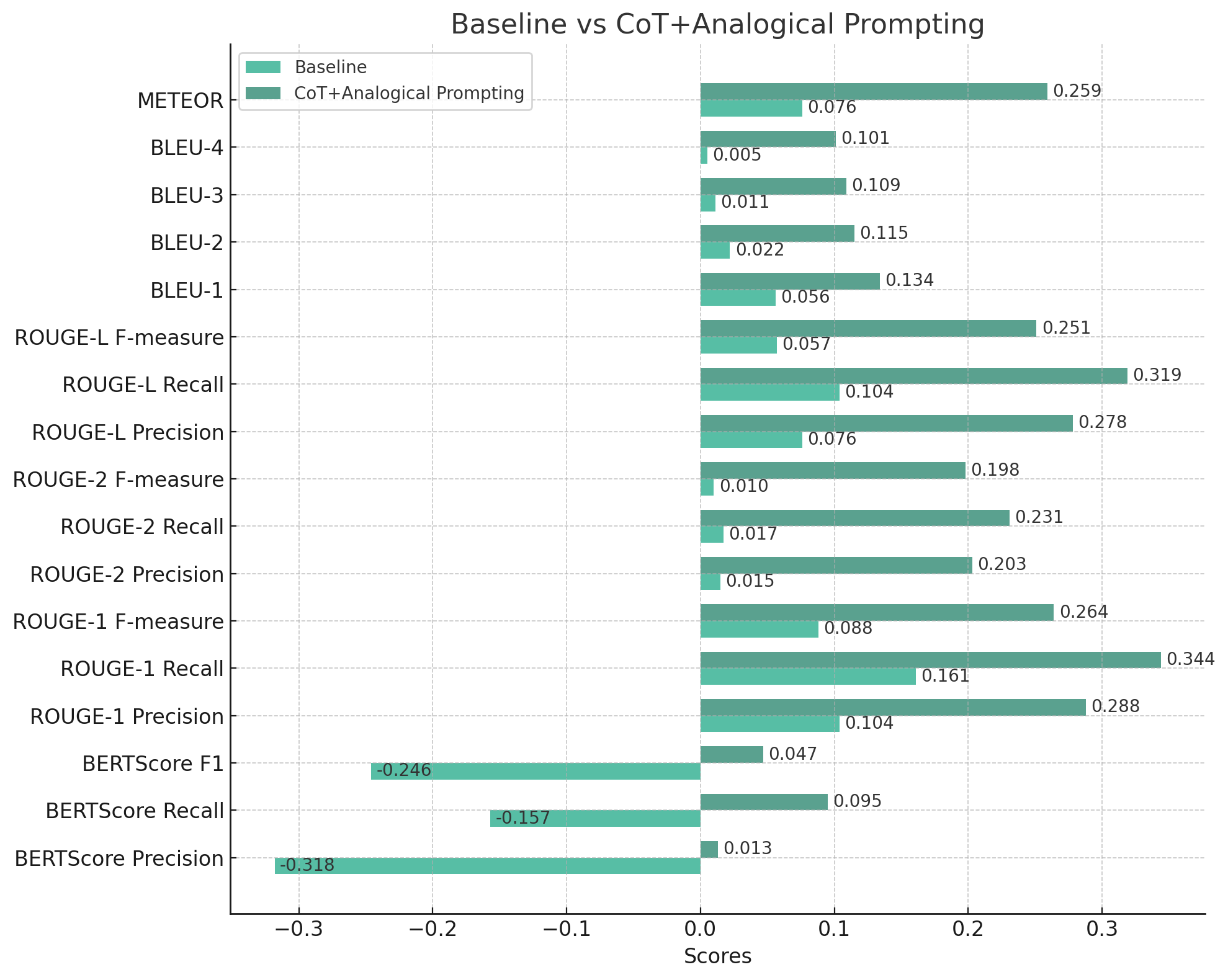} 
  \caption{CoT+Analogical Prompting on Mixtral 8x7B}
  \label{fig:comparison1}
\end{figure}

\begin{figure}[!h]
  \centering
  \includegraphics[width=\linewidth]{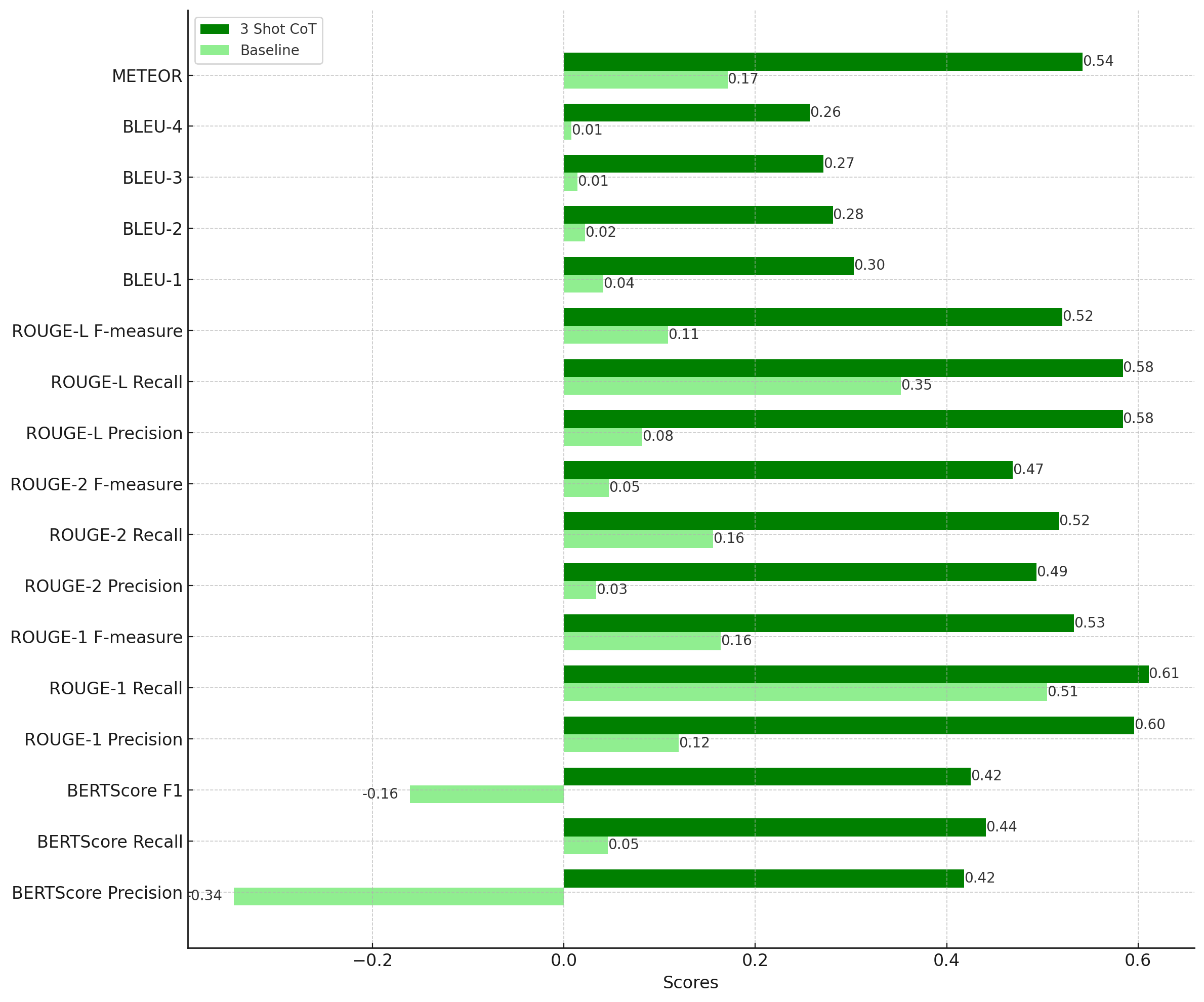} 
  \caption{3 Shot CoT Prompting on Mistral 7B}
  \label{fig:comparison2}
\end{figure}

A detailed evaluation of model outputs was performed to determine the quality of their output. We measured widely recognized metrics in the field of text generation and summarization, including BERTScore\cite{zhang2019bertscore}, METEOR\cite{banerjee2005meteor}, ROUGE-N, and ROUGE-L\cite{lin2004rouge}. 
The \textbf{BERTScores} evaluates how closely the model's generated text aligns with the reference text in terms of content and meaning.
\textbf{METEOR} assesses the congruence between the generated texts and the original answers by integrating precision, recall, and a measure based on the alignment of similar content. In contrast, \textbf{ROUGE-L} focuses on pinpointing the longest sequence of words present in both the generated and the reference solutions, thus offering an indication of the text's smoothness and logical consistency. The \textbf{ROUGE-N} metric expands this evaluation to n-gram co-occurrence analysis, providing a layered dissection of the text's alignment with the ground truth.\newline
We evaluated \textbf{Accuracy} by asking the the evaluator to assess whether they felt the model had generated text inline with the ground truth. We avoided exact matching, as the models are likely to generate text in a different manner than the ground truth.\newline
The results for the experiments can be seen in Tables \ref{tab:accuracy_comparison} and Figures \ref{fig:image1},\ref{fig:comparison},\ref{fig:comparison1},\ref{fig:comparison2}

\begin{table}[h]
\centering
\begin{tabular}{|c|c|}
\hline
\textbf{Model} & \textbf{Baseline Accuracy (\%)} \\
\hline
Mistral-7B      & 31.5  \\
\hline
Mixtral-8x7B      & 42  \\
\hline

\textbf{Model} & \textbf{K-Shot CoT Accuracy (\%)} \\
\hline
Mistral-7B      & 53  \\
\hline
Mixtral-8x7B      & 64.5  \\
\hline

\textbf{Model} & \textbf{Analogical CoT Accuracy (\%)} \\
\hline
Mistral-7B      & 32  \\
\hline
Mixtral-8x7B      & 66.2  \\
\hline
\end{tabular}
\caption{Response Match Accuracy using Human Evaluation}
\label{tab:accuracy_comparison}
\end{table}





\section{Discussion\label{discussion}}
We observe improvements over the baseline upon the application of Analogical CoT prompting in the Mixtral 8x7B model and improvement over the baseline for Mistral 7B on the application of Few Shot CoT.\newline
\textbf{Of note:}
\begin{itemize}
  \item The models became better at giving more structured outputs relative to the baseline
  \item The increase in the recorded metrics points to a better ability to match the meaning and spirit of the ground truth, in this case, the reference answers.
  \item The models were likely held back by a lack of training data specifically in this domain. It is likely that upon finetuning the models on a dataset such as SciQ, JeeBench, or the dataset contributed by \cite{anand2023revolutionizing} 
  , the models would perform better on the given tasks.
\end{itemize}
Additionally, we observed a trend in line with \cite{jin2024impact} with respect to the length of the reasoning chain, i.e, longer (or more complex) questions benefited more from longer steps and a higher number of steps. \newline
We also observed that prompting the model with the steps explicitly delineated led to a better quality answer, along with a slight boost in accuracy relative to the baseline.

\section{Limitations\label{limitations}}
\textbf{Analogical Prompting} is a relatively taxing technique, as it requires the model to be trained on a large amount of domain-specific data in order to be able to recall relevant examples.
However, as noted in the original work, this method of prompting does hold merit for larger, better trained models such as GPT4.
Theoretically, one could expect to see similar performance on open source models, should they be finetuned on specialist data.\newline\newline
\textbf{Mixture of Expert Models} are somewhat difficult to work with. While they do boast faster inference times due to their router layers, the same attribute makes it difficult to finetune these models without overfitting. 4-Bit Quantizing\cite{dettmers2023qlora} these models also leads to a degradation in performance, as the Mixtral model is essentially eight 7B models stacked together with a router layer, and smaller models tend to experience heavier degradation from quantization.\newline\newline
The \textbf{Data Comprehensiveness} of StemStep may be lower than is optimal for finetuning LLMs, as it contains only 928 questions. By expanding this dataset, we may see further utility with respect to prompt construction and evaluation.

\section{Future Scope\label{future scope}}
The \textbf{Mixture of Experts} approach could hypothetically be applied to enhance question-answering capabilities in STEM subjects by leveraging specialized models for different subdomains of physics for example, such as classical mechanics, quantum physics, and thermodynamics. Each expert, knowledgeable about its specific area, can provide precise and nuanced answers. When integrated, this collective expertise allows for a comprehensive and accurate response to diverse and complex physics questions, making MoE a promising framework for tackling interdisciplinary challenges in question answering.
It is feasible that prompt engineering designed to exploit such a model could lead to a high accuracy, high quality Question Answering system in a given domain.\newline
\textbf{Supervised Finetuning} of Mixtral 8x7B may allow the model to better leverage Analogical prompting due to the increased specialist knowledge in the given domains enabling the model to better recall relevant examples to bolster its own 'understanding' of the topic at hand.\newline
Further expansion of the \textbf{StemStep} dataset would improve its utility for prompt evaluation and model training, allowing models to learn the ideal structure for such domain-specific queries through methods such as RLHF\cite{ouyang2022training}.

\section{Conclusion\label{conclusion}}
 Through this work, we present both \textbf{StemStep}, as well as a potential avenue for leveraging analogical prompting on open source models \textbf{for educational purposes}.
\textbf{StemStep} is a new dataset containing 1000 Physics and Mathematics questions, as well as the steps required to solve them. We believe this contribution will help others who wish to evaluate and finetune LLMs in the STEM Education domain.\newline
 Analogical prompting was floated as an alternative to the more resource intensive CoT prompting methods, however, most open source models are unable to leverage it to any real degree of usefulness.\newline
 The \textbf{Analogical Chain of Thought Prompting} method we have proposed could serve to bridge the gap between closed and open source models, although further evaluation on finetuned models is neccesary to draw stronger conclusions.\newline
 We hope to move STEM education towards a new paradigm of\textbf{ Student Centric} topic explanations, as opposed to the current paradigm centered on the ability and willingness of the teacher to impart knowledge, and in doing so, improve knowledge in this domain significantly.\newpage

\begin{appendix}

\section{Appendix: Additional Experimentation Details}
\subsection*{Experimentation}\label{app:experiments}
To evaluate model performance on each prompting method, we prompt the model using questions from \textbf{StemStep} and compare the resulting output with the ground truth present in the dataset.\newline
For example, given a question on the topic of \textbf{kinematics}:\newline

\textbf{Instruction:}  
"A car accelerates from rest at a constant rate of \( 2 \, m/s^2 \). Calculate the time it takes to reach a velocity of \( 20 \, m/s \)."\newline

\textbf{Analogical Prompt:}  \newline
Recall relevant exemplars and solve the question: "A car accelerates from rest at a constant rate of \( 2 \, m/s^2 \). Calculate the time it takes to reach a velocity of \( 20 \, m/s \)."  \newline

Such a prompt would be applied to evaluate Analogical prompting, the generated output of the model would then be compared to the ground truth, using the metrics described in \ref{metrics}, along with fuzzy matching, which was used to regenerate responses which had a similarity of less than \textbf{0.3}, in order to ensure responses which greatly deviated from the relevant topic were discarded.
To compare texts using cosine similarity, we first convert the texts into TF-IDF vectors. Let \( \mathbf{T}_A \) and \( \mathbf{T}_B \) be the TF-IDF vectors for texts \( A \) and \( B \), respectively. The cosine similarity between these vectors is given by:

\[
\text{cosine\_similarity}(\mathbf{T}_A, \mathbf{T}_B) = \frac{\mathbf{T}_A \cdot \mathbf{T}_B}{\|\mathbf{T}_A\| \|\mathbf{T}_B\|}
\]

where:
\begin{itemize}
    \item \( \mathbf{T}_A \cdot \mathbf{T}_B = \sum_{i=1}^{n} \mathbf{T}_{A,i} \mathbf{T}_{B,i} \) is the dot product of the TF-IDF vectors.
    \item \( \|\mathbf{T}_A\| = \sqrt{\sum_{i=1}^{n} (\mathbf{T}_{A,i})^2} \) is the norm (magnitude) of vector \( \mathbf{T}_A \).
    \item \( \|\mathbf{T}_B\| = \sqrt{\sum_{i=1}^{n} (\mathbf{T}_{B,i})^2} \) is the norm (magnitude) of vector \( \mathbf{T}_B \).
    \item \( n \) is the number of unique terms in the texts.
\end{itemize}

To regenerate responses that have a similarity of less than 0.3, we use the following condition:
\[
\text{If } \text{cosine\_similarity}(\mathbf{T}_A, \mathbf{T}_B) < 0.3
\]
\[
 \text{ then regenerate response}
\]
\subsection*{Dataset Preprocessing}
The \textbf{StemStep} dataset was preprocessed to ensure consistency and compatibility with the models used in this study. The following steps were performed:

\begin{enumerate}
    \item All LaTeX equations were converted to plain text using a custom script.
    \item Any special characters or symbols were replaced with their corresponding ASCII representations.
    \item Duplicate or near-duplicate questions were removed from the dataset.
    \item The dataset was split into training, validation, and test sets using an 60/20/20 ratio.
\end{enumerate}

\subsection*{Computing Infrastructure}
All experiments were conducted on the following GPU hardware

\begin{itemize}
    \item GPU: 2 x NVIDIA A100 (40GB)
\end{itemize}

\end{appendix}
\bibliography{acl_latex}




\end{document}